\documentclass[conference]{IEEEtran}
\IEEEoverridecommandlockouts
\usepackage{cite}
\usepackage{amsmath,amssymb,amsfonts}
\usepackage{algorithmic}
\usepackage{hyperref}

\usepackage{graphicx}
\usepackage{textcomp}
\usepackage{xcolor}
\def\BibTeX{{\rm B\kern-.05em{\sc i\kern-.025em b}\kern-.08em
    T\kern-.1667em\lower.7ex\hbox{E}\kern-.125emX}}
\begin{document}

\title{Compressed Models Decompress Race Biases: What Quantized Models Forget for Fair Face Recognition
}

\author{\IEEEauthorblockN{Pedro C. Neto\textsuperscript{1,2,*},  Eduarda Caldeira\textsuperscript{1,2}, Jaime S. Cardoso\textsuperscript{1,2} and Ana F. Sequeira\textsuperscript{1}}
\IEEEauthorblockA{\textsuperscript{1}\textit{Centre for Telecommunications and Multimedia, INESC TEC, Porto, Portugal} \\
\textsuperscript{2}\textit{Faculdade de Engenharia da Universidade do Porto, Porto, Portugal}\\
* pedro.d.carneiro@inesctec.pt
}
}
\maketitle

\begin{abstract}
With the ever-growing complexity of deep learning models for face recognition, it becomes hard to deploy these systems in real life. Researchers have two options: 1) use smaller models; 2) compress their current models. Since the usage of smaller models might lead to concerning biases, compression gains relevance. However, compressing might be also responsible for an increase in the bias of the final model. We investigate the overall performance, the performance on each ethnicity subgroup and the racial bias of a State-of-the-Art quantization approach when used with synthetic and real data. This analysis provides a few more details on potential benefits of performing quantization with synthetic data, for instance, the reduction of biases on the majority of test scenarios. We tested five distinct architectures and three different training datasets. The models were evaluated on a fourth dataset which was collected to infer and compare the performance of face recognition models on different ethnicity. 

\end{abstract}

Face recognition methods have made significant progress over the previous years~\cite{boutros2022elasticface}. Current systems are capable of rivalling with humans under certain conditions and are quickly reducing the gap on the remaining test scenarios~\cite{phillips2018face}. The urge to keep the current rate of improvement on these deep learning-based approaches led to an era of complex and obscure models. As such, despite their extraordinary performance, there are two pressing concerns. First, there are hardware limitations that affect the complexity of the models that can be deployed and used in real scenarios. These limitations affect both storage, memory and processing time. The second concern is that the behaviour of a deep neural network is not easily understood~\cite{neto2022explainable}. As such, besides the valuable information, also irrelevant or even harmful correlations can be learnt by these models, and hidden within their obscure nature. 

Addressing these two concerns is of utter importance. To tackle them individually, one must be careful to avoid a potential trade-off between their mitigation. For instance, considering the possibility of an existing bias on the models, further reducing the model size can lead to an increased bias. Moreover, unless that the model is reduced to an interpretable version of itself, these growing biases remain hidden within the black-box model.

Instead of using a smaller model, current work is investigating different model compression approaches. Quantization, knowledge distillation and pruning are the most common. In this work, we aim to study the impact of quantization on the mitigation or amplification of existing biases. Hooker~\textit{et al.}~\cite{hooker2020characterising} presented a set of experiments that indicates a potential increase of the previous biases and tried to identify the elements forgotten by the deep neural network~\cite{hooker2019compressed}. To further extend this research, we framed our problem within the context of racial biases in face recognition systems. Stoychev~\textit{et al.}~\cite{stoychev2022effect} presented mixed results on a face-related task and the effects on the biases were dependent on the training dataset. For this reason, our work, starting from Boutros~\textit{et al.}~\cite{boutros2022quantface} quantization approach, further includes the usage of real and synthetic data and the usage of distinct datasets for training. This study also aims to understand the current trade-offs between small models and hidden biases. 

\begin{figure*}
    \centering
    \includegraphics[width=\linewidth]{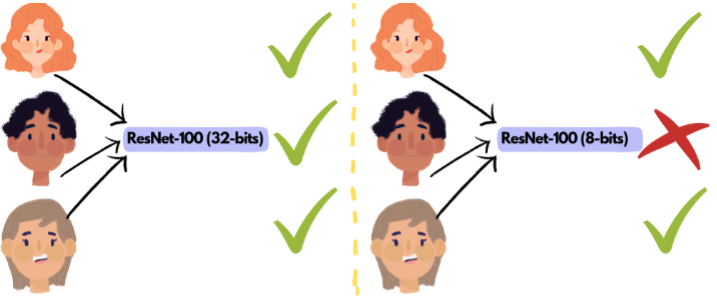}
    \caption{Representation on a potential source of biases created after quantizing a deep neural network. Less represented classes or classes prone to suffer from discrimination might be easier to forget, leading to errors focused on those classes. }
    \label{fig:enter-label}
\end{figure*}

Within the context of this work, we aim to answer three research questions: 1) \textit{Are smaller models more biased?} 2) \textit{Are quantized models more biased? (As represented in Figure~\ref{fig:enter-label}}) 3) \textit{What is the impact of using synthetic data to quantize these models}? We individually address each of these questions and provide a potential explanation for the behaviour displayed by the models. Furthermore, the usage of synthetic data is motivated by the possibility of further growing our current datasets without compromising ethical concerns or privacy. Besides, with synthetic data, it is possible to create a higher degree of variability. Given these research questions, we present the following contributions: 
\begin{itemize}
    \item A study on racial bias on four differently sized models trained on MS1MV2~\cite{deng2019arcface} and of two models trained on BUPT-Balancedface and BUPT-Globalface~\cite{wang2021meta};
    \item Using QuantFace~\cite{boutros2022quantface} and real data we study the racial bias of the quantized version of all the models previously evaluated;
    \item We repeat the previous analysis with synthetic data;
    \item While confirming some of the Hooker~\textit{et al.}~\cite{hooker2020characterising} findings, we have also discovered that quantization with synthetic data mitigates the racial bias of the resulting model. 
\end{itemize}
While we conducted our studies on a face recognition task, we theorise that the results seen are plausible for similar verification or classification problems, and other biometric modalities. In other words, smaller models and quantisation might impact the class with higher risk of discrimination in the same way that we describe in this study.  

The following sections are divided into four major sections and a conclusion. Section~\ref{sec:dataset} describes the five distinct datasets utilised in this study. Afterwards, in Section~\ref{sec:methods} and \ref{sec:setuo} the methodology and experimental setup are described in detail. Finally, the results are shown and discussed in Section~\ref{sec:reults}.

\section{Datasets}
\label{sec:dataset}

This study utilized five distinct datasets. MS1MV2, BUPT-Balancedface and BUPT-Globalface have been used for training the base models and quantization, while the synthetic data was only used for quantization and RFW just for evaluation of the models. 

\subsection{MS1MV2}

MS1MV2 is widely used in the literature to train and compare several deep face recognition models~\cite{boutros2022elasticface,neto2021focusface}. It is a refined version of the original MS-Celeb-1M dataset~\cite{guo2016ms}, which further improved the training of these systems. The dataset contains 85k different identities and almost six million images and it is not balanced with respect to the race. 

\subsection{BUPT-Balancedface and BUPT-Globalface}

Wang~\textit{et al.}~\cite{wang2021meta} introduced two distinct datasets to train deep face recognition systems. These datasets have been created to mitigate race bias on face recognition through skin tone labelling as African, Asian, Caucasian and Indian. BUPT-Globalface contains two million images from 38k different identities, and the distribution of races follows their distribution in the world. On the other hand, BUPT-Balancedface contains 1.3 million images from 28k identities which are divided into 7k identities per race. As such, this second dataset is race balanced.

\subsection{Synthetic data}

This dataset, introduced in~\cite{boutros2022quantface} contains approximately 500k unlabelled synthetic images. These images have been generated by a generative adversarial network~\cite{gan,DBLP:conf/nips/KarrasAHLLA20}. The noise used as input to generate the images was sampled from a Gaussian distribution and fed to a pretrained generator (official open source implementation \footnote{\url{https://github.com/NVlabs/stylegan2-ada}} of StyleGAN2-AD). The usage of synthetic data is often seen to result in sub-optimal performances \cite{Qiu_2021_ICCV} which might be caused by a domain gap between real and synthetic data~\cite{DBLP:conf/aaai/XuZNLWTZ20,DBLP:conf/cvpr/Sankaranarayanan18,DBLP:conf/cvpr/LeePYL20}. In this work, the goal is not to use the synthetic data to learn the representations from scratch, and we further argue that there might exist advantages of this domain gap. 

\subsection{RFW}

Proposed by the same authors of BUPT-Balancedface, Racial Faces in-the-wild (RFW)~\cite{Wang_2019_ICCV} was designed as a benchmarking dataset for fair face verification. Similarly, it includes labels for ethnicity, which allows for a fair assessment of potential biases. It contains 3000 individuals with 6000 image pairs for face verification.

\section{Methods}
\label{sec:methods}

The methodology was designed in two different processes. First, it is necessary to understand if there is a bias problem on quantized models, and for this we have used the publicly available QuantFace models. If the problem is identified, it is necessary to understand if it is visible on models trained on other datasets (balanced and non-balanced). 

QuantFace has four different architectures available: MobileFaceNet~\cite{mobilefacenet}, ResNet-18~\cite{resnet}, ResNet-50 and ResNet-100. Each of these architectures is available in five distinct shapes: the original full-precision model, the 8-bit model quantized with real data, the 8-bit model quantized with synthetic data, the 6-bit model quantized with real data and finally the 6-bit model quantized with synthetic data. This part is essential to understand if the behaviour of the quantized model changes with the selected precision and the network architecture. Hence, the dataset for this is fixed as the MS1MV2, so we can ignore the data as a factor of variability. 

For the second part of the study, a ResNet-34 was trained on BUPT-Balancedface and a second ResNet-34 was trained on BUPT-Globalface. The first is available in the four different quantized versions described above, whereas the second was only studied in its 8-bit version with real data quantization from two different sources and synthetic quantization. The network architecture is fixed so that the variability factors are limited to the data used for training and quantization. Moreover, the usage of a different real dataset for quantization than the one used for training attempts to further improve the understanding of the reasons behind the performance of a model. For instance, performance changes might be caused by the usage of data from a different distribution than the training data. 

In order to gain additional insights regarding the reasons behind the impact of the synthetic data on the bias resulting from the quantization, we further trained a ethnicity classifier on BUPT-Balancedface to estimate the ethnicity distribution in the synthetic data. This classifier comprises a fully-connect layer on top of a pretrained Elastic-Arc model~\cite{boutros2022elasticface} model and achieves accuracies above 95\%.

\section{Experimental Setup}
\label{sec:setuo}

For the quantization process we have utilized the open-source implementation of QuantFace\footnote{\url{https://github.com/fdbtrs/QuantFace/}} with a batch size of 128 on a Nvidia Tesla v100 32GB. We have utilized the same configuration as proposed by Boutros~\textit{et al.}~\cite{boutros2022quantface}. For training the face recognition models on BUPT-Balancedface and BUPT-Globalface, we have utilized the same protocol of Deng~\textit{et al.}~\cite{deng2019arcface} to preprocess the images, reduce the learning rate and stop the training. The ethnicity classifier utilised the Elastic-Arc model~\cite{boutros2022elasticface} with all its layers frozen. An additional classifier layer was added and trained.

\subsection{Evaluation Metrics}

    The performance of the evaluated models was measures in terms of accuracy. For the fairness evaluation of these models we have utilised two metrics: the standard deviation between the different accuracies (STD), and the skewed error ratio (SER) seen in Equation~\ref{eq:ser}. 

\begin{equation}
    SER = \frac{100-min(acc)}{100-max(acc)}
    \label{eq:ser}
\end{equation}

    The STD aims to evaluate the variance between the different accuracy values. The usage of STD in a set with just four different samples might be questioned, however, this has been the approach used in the literature~\cite{wang2021meta}. For the sake of reproducible research and compliance with the literature, we have chosen to retain the metrics previously used. On the other hand, SER measures or much larger is the worst error when compared with the better error. This is important to understand the relative differences between the different accuracy values. As a relative evaluation metric, SER is highly sensitive when the accuracy is above 99\%. This happens because as the errors get below 1\% their relative difference also change accordingly. For instance, a SER computed for a maximum accuracy of 90\% and a minimum accuracy of 80\% is the same if these accuracy values were 99.9\% and 99.8\%. STD is highly sensitive to absolute differences, and grows large on sets with lower accuracy values.  

\section{Results}
\label{sec:reults}

A careful analysis of the performance of the different sized models at full precision (Table~\ref{tab:results_ms1mv2}) shows that smaller models tend to have higher biases and lower performance in terms of average accuracy. ResNet-100 is an exception and this difference might be connected to the fact that SER becomes highly sensitive when the errors are below 1\%. 

\begin{table*}
\normalsize
    \centering
    \caption{Table comprising the results, evaluated on RFW,  from the different models trained on MS1MV2 and their respective quantized versions for different bits and quantization strategies (real or synthetic data). The versions of the models quantized with synthetic data seem to display better fairness metrics at a comparable average performance. }
    \label{tab:results_ms1mv2}
    \begin{tabular}{l|l|l|llll|l|l|l}
    \hline
        Model & Bits & Quant.& Caucasian & Indian & Asian & African  & Avg. & STD & SER  \\ \hline
         & 32 & - & 95.18\% & 92.00\% & 89.93\% & 90.22\% & \textbf{91.83\%} & 2.41 & 2.09  \\ 
         & 8& Real & 95.32\% & 91.60\% & 89.27\% & 90.08\% & 91.57\% & 2.68 & 2.29  \\
         MobileFaceNets & 8& Synth. & 94.18\% & 91.83\% & 88.85\% & 89.72\% & 91.15\% &\textbf{ 2.38} & 1.92  \\ 
         & 6 & Real & 90.05\% & 86.52\% & 82.88\% & 83.18\% & 85.66\% & 3.36 & 1.72  \\ 
         & 6 & Synth. & 89.97\% & 86.95\% & 83.13\% & 84.40\% & 86.11\% & 3.02 & \textbf{1.68} \\ \hline
         & 32 & - & 97.48\% & 95.38\% & 93.72\% & 94.27\% & \textbf{95.21\%} & 1.66 & 2.49  \\ 
         & 8& Real &97.42\% & 95.33\% & 93.55\% & 94.20\% & 95.13\% & 1.70 & 2.50 \\ 
        ResNet-18 & 8& Synth. &96.95\% & 95.07\% & 93.30\% & 93.87\% & 94.80\% & \textbf{1.61} & \textbf{2.20}  \\ 
         & 6 & Real &96.93\% & 94.65\% & 92.52\% & 93.22\% & 94.33\% & 1.95 & 2.44  \\ 
         & 6 & Synth. &96.80\% & 94.78\% & 92.35\% & 93.28\% & 94.30\% & 1.94 & 2.39  \\ \hline
         & 32 & - & 99.00\% & 98.15\% & 97.62\% & 98.32\% & 98.27\% & \textbf{0.57} & \textbf{2.38 }\\ 
         & 8& Real & 99.07\% & 98.07\% & 97.65\% & 98.40\% & \textbf{98.30\%} & 0.60 & 2.53 \\
         ResNet-50 & 8& Synth. & 99.02\% & 97.72\% & 97.33\% & 97.88\% & 97.99\% & 0.73 & 2.72  \\
         & 6 & Real & 98.32\% & 96.27\% & 94.55\% & 95.87\% & 96.25\% & 1.56 & 3.24  \\
         & 6 & Synth. & 97.95\% & 96.63\% & 94.97\% & 96.20\% & 96.44\% & 1.23 & 2.45  \\ \hline
         & 32 & - &  99.65\% & 98.88\% & 98.50\% & 99.00\% & \textbf{99.01\%} & \textbf{0.48} & 4.29 \\ 
         & 8& Real &  99.57\% & 98.87\% & 98.15\% & 98.77\% & 98.84\% & 0.58 & 4.30  \\
         ResNet-100 & 8& Synth. & 99.37\% & 98.72\% & 98.13\% & 98.78\% & 98.75\% & 0.51 & 2.97  \\
         & 6 & Real &  95.27\% & 93.15\% & 90.32\% & 91.70\% & 92.61\% & 2.12 & 2.05 \\
         & 6 & Synth. & 95.93\% & 93.40\% & 91.92\% & 92.60\% & 93.46\% & 1.75 & \textbf{1.99} \\ \hline
        
    \end{tabular}
\end{table*}

The quantized version of these models seems to retain the performance and bias advantaged when compared to simpler models. As theorised, the quantization has a negative impact on the bias, and in most cases on the performance too. The lower the number of bit, the higher the bias. However, the usage of synthetic data has shown, for all the different precisions, a capability to reduce the bias while retaining the performance. From this data, it is not clear if the improvement is due to a specific characteristic of the synthetic data. 

We have used the ethnicity classifier to get an estimation of the racial balance of the synthetic data. We obtained 365889 Caucasians, 81568 Asians, 81568 Indians and 61966 Africans. Since the data is not balanced, it is not possible to associate the effects of this data to its balance.

\begin{table*}
\normalsize
    \centering
    \caption{Table comprising the results, evaluated on RFW, from two ResNet-34 models trained on BUPT-Balancedface (BL) and BUPT-Globalface (GL) and their respective quantized versions for different bits and quantization strategies (BL, GL or synthetic data). The versions of the models quantized with synthetic data seem to perform outstandingly well. }
    \label{tab:results_balance}
    \begin{tabular}{l|l|l|llll|l|l|l}
    \hline
        Train Data & Bits & Quant.& Caucasian & Indian & Asian & African  & Avg. & STD & SER \\ \hline
         & 32 & - & 96.60\% & 94.50\% & 94.03\% & 93.37\% & \textbf{94.63\%} & \textbf{1.40} & 1.95 \\ 
         & 8& BL & 94.98\% & 93.60\% & 92.77\% & 90.95\% & 93.08\% & 1.68 & 1.80  \\
         BL & 8& Synth. & 96.03\% & 94.40\% & 93.97\% & 92.50\% & 94.23\% & 1.45 & 1.89  \\ 
         & 6 & BL &  89.22\% & 87.87\% & 86.25\% & 82.80\% & 86.54\% & 2.77 & \textbf{1.60}  \\ 
         & 6 & Synth. & 94.58\% & 92.88\% & 91.45\% & 91.13\% & 92.51\% & 1.58 & 1.64 \\ \hline
         & 32 & - & 97.67\% & 95.52\% & 94.15\% & 93.87\% & \textbf{95.30\%} & 1.74 & 2.63  \\ 
         & 8& BL & 95.42\% & 92.75\% & 91.83\% & 89.88\% & 92.47\% & 2.30 & 2.21 \\ 
        GL & 8& GL. & 94.70\% & 92.15\% & 90.23\% & 88.75\% & 91.46\% & 2.57 & \textbf{2.12}  \\ 
         & 8 & Synth. & 97.33\% & 95.15\% & 94.17\% & 93.55\% & 95.05\% & \textbf{1.66} & 2.42 \\ \hline

    \end{tabular}
\end{table*}

Further training two ResNet-34 on BUPT-Balancedface and BUPT-Globalface shows, at full precision, that despite a higher performance of the latter, the balance of the former is essential to ensure better bias metrics. On the model trained with the BUPT-Balancedface the versions quantized with synthetic data has not only kept the same tendency of the previous table, but it has also surpassed by a large margin the version of the method quantized with the balanced data. This might be caused by the lower variability of the BUPT-Balancedface data with respect to the number of identities. 

The ResNet-34 trained on the BUPT-Globalface performs better if quantized with the data from the BUPT-Balancedface instead of using the data from training. Once again, it might be possible that introducing variability and unseen data for the quantization increases the capability of the model to be robust for all ethnicities. This is further validated by the version quantized with the synthetic data, which leads to a performance similar to the full precision model. 

\section{Conclusion}
\label{sec:conclusion}

In this document, we tackled three research questions, and we have provided answers to all of them. 1) and 2) It was possible to infer that models quantized with real data and smaller models are indeed more biased;  3) it was also verifiable that using synthetic data for quantization positively impacts the fairness metrics. We have extended previous literature on the assessment of the information that is lost by quantized models and further introduced a novel topic regarding the usage of synthetic data for bias mitigation. 

Despite the interesting results shown by our experiments, there are several gaps in the literature that should be tackled in future work. For instance, it is not known if this behaviour is the same for gender biases, or if synthetic data harms gender biases while helping to mitigate race biases. A more comprehensive study is required. The usage of the combined real data that has and has not been seen, and synthetic data should be also analysed to understand how can we, just by changing the training data, mitigate these biases while retaining the original performance.  Furthermore, we still do not know if these findings hold for different traits and tasks, and further studies are required to confirm the generalization of these findings. 

While the results shown are still preliminary, they introduce a few research directions that might be relevant for the future of biometrics in an era of increasing concern with these biases.

\section*{Acknowledgments}

This work is co-financed by Component 5 - Capitalization and Business Innovation, integrated in the Resilience Dimension of the Recovery and Resilience Plan within the scope of the Recovery and Resilience Mechanism (MRR) of the European Union (EU), framed in the Next Generation EU, for the period 2021 - 2026, within project NewSpacePortugal, with reference 11. It was also financed by National Funds through the Portuguese funding agency, FCT - Fundação para a Ciência e a Tecnologia within the PhD grant ``2021.06872.BD''. 

\bibliographystyle{IEEEtran}
\bibliography{ref}

\begin{thebibliography}{10}
\providecommand{\url}[1]{#1}
\csname url@samestyle\endcsname
\providecommand{\newblock}{\relax}
\providecommand{\bibinfo}[2]{#2}
\providecommand{\BIBentrySTDinterwordspacing}{\spaceskip=0pt\relax}
\providecommand{\BIBentryALTinterwordstretchfactor}{4}
\providecommand{\BIBentryALTinterwordspacing}{\spaceskip=\fontdimen2\font plus
\BIBentryALTinterwordstretchfactor\fontdimen3\font minus
  \fontdimen4\font\relax}
\providecommand{\BIBforeignlanguage}[2]{{%
\expandafter\ifx\csname l@#1\endcsname\relax
\typeout{** WARNING: IEEEtran.bst: No hyphenation pattern has been}%
\typeout{** loaded for the language `#1'. Using the pattern for}%
\typeout{** the default language instead.}%
\else
\language=\csname l@#1\endcsname
\fi
#2}}
\providecommand{\BIBdecl}{\relax}
\BIBdecl

\bibitem{boutros2022elasticface}
F.~Boutros, N.~Damer, F.~Kirchbuchner, and A.~Kuijper, ``Elasticface: Elastic
  margin loss for deep face recognition,'' in \emph{Proceedings of the IEEE/CVF
  conference on computer vision and pattern recognition}, 2022, pp. 1578--1587.

\bibitem{phillips2018face}
P.~J. Phillips, A.~N. Yates, Y.~Hu, C.~A. Hahn, E.~Noyes, K.~Jackson, J.~G.
  Cavazos, G.~Jeckeln, R.~Ranjan, S.~Sankaranarayanan \emph{et~al.}, ``Face
  recognition accuracy of forensic examiners, superrecognizers, and face
  recognition algorithms,'' \emph{Proceedings of the National Academy of
  Sciences}, vol. 115, no.~24, pp. 6171--6176, 2018.

\bibitem{neto2022explainable}
P.~C. Neto, T.~Gon{\c{c}}alves, J.~R. Pinto, W.~Silva, A.~F. Sequeira, A.~Ross,
  and J.~S. Cardoso, ``Explainable biometrics in the age of deep learning,''
  \emph{arXiv preprint arXiv:2208.09500}, 2022.

\bibitem{hooker2020characterising}
S.~Hooker, N.~Moorosi, G.~Clark, S.~Bengio, and E.~Denton, ``Characterising
  bias in compressed models,'' \emph{arXiv preprint arXiv:2010.03058}, 2020.

\bibitem{hooker2019compressed}
S.~Hooker, A.~Courville, G.~Clark, Y.~Dauphin, and A.~Frome, ``What do
  compressed deep neural networks forget?'' \emph{arXiv preprint
  arXiv:1911.05248}, 2019.

\bibitem{stoychev2022effect}
S.~Stoychev and H.~Gunes, ``The effect of model compression on fairness in
  facial expression recognition,'' \emph{arXiv preprint arXiv:2201.01709},
  2022.

\bibitem{boutros2022quantface}
F.~Boutros, N.~Damer, and A.~Kuijper, ``Quantface: Towards lightweight face
  recognition by synthetic data low-bit quantization,'' in \emph{2022 26th
  International Conference on Pattern Recognition (ICPR)}.\hskip 1em plus 0.5em
  minus 0.4em\relax IEEE, 2022, pp. 855--862.

\bibitem{deng2019arcface}
J.~Deng, J.~Guo, N.~Xue, and S.~Zafeiriou, ``Arcface: Additive angular margin
  loss for deep face recognition,'' in \emph{{IEEE} Conference on Computer
  Vision and Pattern Recognition, {CVPR} 2019, Long Beach, CA, USA, June 16-20,
  2019}.\hskip 1em plus 0.5em minus 0.4em\relax Computer Vision Foundation /
  {IEEE}, 2019, pp. 4690--4699.

\bibitem{wang2021meta}
M.~Wang, Y.~Zhang, and W.~Deng, ``Meta balanced network for fair face
  recognition,'' \emph{IEEE Transactions on Pattern Analysis and Machine
  Intelligence}, 2021.

\bibitem{neto2021focusface}
P.~C. Neto, F.~Boutros, J.~R. Pinto, N.~Damer, A.~F. Sequeira, and J.~S.
  Cardoso, ``Focusface: Multi-task contrastive learning for masked face
  recognition,'' in \emph{2021 16th IEEE International Conference on Automatic
  Face and Gesture Recognition (FG 2021)}.\hskip 1em plus 0.5em minus
  0.4em\relax IEEE, 2021, pp. 01--08.

\bibitem{guo2016ms}
Y.~Guo, L.~Zhang, Y.~Hu, X.~He, and J.~Gao, ``Ms-celeb-1m: A dataset and
  benchmark for large-scale face recognition,'' in \emph{Computer Vision--ECCV
  2016: 14th European Conference, Amsterdam, The Netherlands, October 11-14,
  2016, Proceedings, Part III 14}.\hskip 1em plus 0.5em minus 0.4em\relax
  Springer, 2016, pp. 87--102.

\bibitem{gan}
I.~J. Goodfellow, J.~Pouget{-}Abadie, M.~Mirza, B.~Xu, D.~Warde{-}Farley,
  S.~Ozair, A.~C. Courville, and Y.~Bengio, ``Generative adversarial nets,'' in
  \emph{Advances in Neural Information Processing Systems 27: Annual Conference
  on Neural Information Processing Systems 2014, December 8-13 2014, Montreal,
  Quebec, Canada}, 2014, pp. 2672--2680.

\bibitem{DBLP:conf/nips/KarrasAHLLA20}
T.~Karras, M.~Aittala, J.~Hellsten, S.~Laine, J.~Lehtinen, and T.~Aila,
  ``Training generative adversarial networks with limited data,'' in
  \emph{Advances in Neural Information Processing Systems 33: Annual Conference
  on Neural Information Processing Systems 2020, NeurIPS 2020, December 6-12,
  2020, virtual}, 2020.

\bibitem{Qiu_2021_ICCV}
H.~Qiu, B.~Yu, D.~Gong, Z.~Li, W.~Liu, and D.~Tao, ``Synface: Face recognition
  with synthetic data,'' in \emph{Proceedings of the IEEE/CVF International
  Conference on Computer Vision (ICCV)}, October 2021, pp. 10\,880--10\,890.

\bibitem{DBLP:conf/aaai/XuZNLWTZ20}
M.~Xu, J.~Zhang, B.~Ni, T.~Li, C.~Wang, Q.~Tian, and W.~Zhang, ``Adversarial
  domain adaptation with domain mixup,'' in \emph{The Thirty-Fourth {AAAI}
  Conference on Artificial Intelligence, {AAAI} 2020, The Thirty-Second
  Innovative Applications of Artificial Intelligence Conference, {IAAI} 2020,
  The Tenth {AAAI} Symposium on Educational Advances in Artificial
  Intelligence, {EAAI} 2020, New York, NY, USA, February 7-12, 2020}.\hskip 1em
  plus 0.5em minus 0.4em\relax {AAAI} Press, 2020, pp. 6502--6509.

\bibitem{DBLP:conf/cvpr/Sankaranarayanan18}
S.~Sankaranarayanan, Y.~Balaji, A.~Jain, S.~Lim, and R.~Chellappa, ``Learning
  from synthetic data: Addressing domain shift for semantic segmentation,'' in
  \emph{2018 {IEEE} Conference on Computer Vision and Pattern Recognition,
  {CVPR} 2018, Salt Lake City, UT, USA, June 18-22, 2018}.\hskip 1em plus 0.5em
  minus 0.4em\relax Computer Vision Foundation / {IEEE} Computer Society, 2018,
  pp. 3752--3761.

\bibitem{DBLP:conf/cvpr/LeePYL20}
S.~Lee, E.~Park, H.~Yi, and S.~H. Lee, ``Strdan: Synthetic-to-real domain
  adaptation network for vehicle re-identification,'' in \emph{2020 {IEEE/CVF}
  Conference on Computer Vision and Pattern Recognition, {CVPR} Workshops 2020,
  Seattle, WA, USA, June 14-19, 2020}.\hskip 1em plus 0.5em minus 0.4em\relax
  Computer Vision Foundation / {IEEE}, 2020, pp. 2590--2597.

\bibitem{Wang_2019_ICCV}
M.~Wang, W.~Deng, J.~Hu, X.~Tao, and Y.~Huang, ``Racial faces in the wild:
  Reducing racial bias by information maximization adaptation network,'' in
  \emph{The IEEE International Conference on Computer Vision (ICCV)}, October
  2019.

\bibitem{mobilefacenet}
\BIBentryALTinterwordspacing
S.~Chen, Y.~Liu, X.~Gao, and Z.~Han, ``Mobilefacenets: Efficient cnns for
  accurate real-time face verification on mobile devices,'' in \emph{CCBR 2018,
  Urumqi, China, August 11-12, 2018, Proceedings}, ser. Lecture Notes in
  Computer Science, vol. 10996.\hskip 1em plus 0.5em minus 0.4em\relax
  Springer, 2018, pp. 428--438. [Online]. Available:
  \url{https://doi.org/10.1007/978-3-319-97909-0\_46}
\BIBentrySTDinterwordspacing

\bibitem{resnet}
\BIBentryALTinterwordspacing
K.~He, X.~Zhang, S.~Ren, and J.~Sun, ``Deep residual learning for image
  recognition,'' in \emph{2016 {IEEE} Conference on Computer Vision and Pattern
  Recognition, {CVPR} 2016, Las Vegas, NV, USA, June 27-30, 2016}, 2016, pp.
  770--778. [Online]. Available: \url{https://doi.org/10.1109/CVPR.2016.90}
\BIBentrySTDinterwordspacing

\end{thebibliography}
\end{document}